%% file: GandhiCechHoraud-ICRA2012.tex
\documentclass[a4paper, 10pt, conference]{ieeeconf}      % Use this line for a4
\IEEEoverridecommandlockouts                              % This command is only
\usepackage{graphicx} % for pdf, bitmapped graphics files
\usepackage{subfigure}
\usepackage{amsmath} % assumes amsmath package installed
\usepackage{amssymb}  % assumes amsmath package installed
\usepackage[ruled]{algorithm}
\usepackage {algorithmic}
\usepackage{color}

\title{\LARGE \bf
High-Resolution Depth Maps Based on TOF-Stereo Fusion
}

\author{Vineet Gandhi$^{\dag}$\thanks{${\dag}$ V. Gandhi acknowledges support from the Erasmus Mundus CIMET master program.}, Jan {\v C}ech, and Radu Horaud \\% <-this % stops a space
Inria Grenoble Rh\^{o}ne-Alpes, Montbonnot Saint-Martin, France \\
\{vineet.gandhi, jan.cech, radu.horaud\}@inria.fr \\
}

\DeclareMathSizes{10}{9}{6}{8}

\def\simil{\operatorname{simil}}
\def\argmax{\operatorname{argmax}}

\begin{document}

\maketitle
\thispagestyle{empty}
\pagestyle{empty}

%%%%%%%%%%%%%%%%%%%%%%%%%%%%%%%%%%%%%%%%%%%%%%%%%%%%%%%%%%%%%%%%%%%%%%%%%%%%%%%%
\begin{abstract}
\input{abstract}

\end{abstract}

%%%%%%%%%%%%%%%%%%%%%%%%%%%%%%%%%%%%%%%%%%%%%%%%%%%%%%%%%%%%%%%%%%%%%%%%%%%%%%%%
\section{Introduction}
\input{introduction}

% Problem statement beginning

%%%%%%%%%%%%%%%%%%%%%%%%%%%%%%%%%%%%%%%%%%%%%%%%%%%%%%%%%%%%%%%%%%%%%%%%%%%%%%%%
\section{The proposed algorithm} \label{section:method}

\input{method}

%%%%%%%%%%%%%%%%%%%%%%%%%%%%%%%%%%%%%%%%%%%%%%%%%%%%%%%%%%%%%%%%%%%%%%%%%%%%%%%%
\section{Experiments} \label{section:evaluation}

\input{experiments}
\section{Conclusions} \label{section:conclusions}

\input{conclusion}

%%%%%%%%%%%%%%%%%%%%%%%%%%%%%%%%%%%%%%%%%%%%%%%%%%%%%%%%%%%%%%%%%%%%%%%%%%%%%%%%
%\section{ACKNOWLEDGMENTS}

%\green{This work was part of master thesis of Vineet Gandhi and he was supported by the Erasmus Mundus grant under CIMET consortium.}

%%%%%%%%%%%%%%%%%%%%%%%%%%%%%%%%%%%%%%%%%%%%%%%%%%%%%%%%%%%%%%%%%%%%%%%%%%%%%%%%

\bibliographystyle{IEEEtranS}
%\bibliography{tof_fusion}

\end{document}

%% file: abstract.tex
The combination of range sensors with color cameras can be very useful for robot navigation, semantic perception, manipulation, and telepresence. Several methods of combining range- and color-data have been investigated and successfully used in various robotic applications. Most of these systems suffer from the problems of noise in the range-data and resolution mismatch between the range sensor and the color cameras, since the resolution of current range sensors is much less than the resolution of color cameras. High-resolution depth maps can be obtained using  stereo matching, but this often fails to construct accurate depth maps of weakly/repetitively textured scenes, or if the scene exhibits complex self-occlusions. Range sensors provide coarse depth information regardless of presence/absence of texture. The use of a calibrated system, composed of a time-of-flight (TOF) camera and of a stereoscopic camera pair, allows data fusion thus overcoming the weaknesses of both individual sensors. We propose a novel TOF-stereo fusion method based on an efficient seed-growing algorithm which uses the TOF data projected onto the stereo image pair as an  initial set of correspondences. These initial ``seeds'' are then propagated based on a Bayesian model which combines an image similarity score with rough depth priors computed from the low-resolution range data. The overall result is a dense and accurate depth map at the resolution of the color cameras at hand. We show that the proposed algorithm outperforms 2D image-based stereo algorithms and that the results are of higher resolution than off-the-shelf color-range sensors, e.g., Kinect.
Moreover, the algorithm potentially exhibits real-time performance on a single~CPU.

%% file: introduction.tex
An advanced computer vision system should be able to provide both accurate \textit{color and depth} information for each pixel at high resolution. Such a system can be very useful for automated vision problems especially in the context of robotics, e.g., for building dense 3D maps of indoor environments.

The 3D structure of a scene can be reconstructed from two or more 2D views via a \emph{parallax} between corresponding image points. However, it is difficult to obtain accurate pixel-to-pixel matches for scenes of objects without textured surfaces, with repetitive pattern, or in the presence of occlusions. The main drawback is that stereo matching algorithms frequently fail to reconstruct indoor scenes composed of untextured surfaces, e.g., walls, repetitive patterns and surface discontinuities, which are typical in man-made environments. %and in the presence of surface discontinuities, e.g., occluding contours.

% Our approach{

\begin{figure}%[t]
  \centering
  \subfigure[A TOF-stereo setup]{\includegraphics[width=0.7\linewidth]{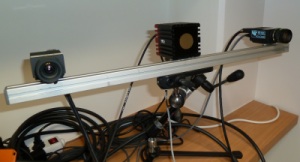}} %\vspace{3.0mm}
  \subfigure[The TOF image is shown in the upper-left corner of a color image.]{\includegraphics[height=0.36\linewidth]{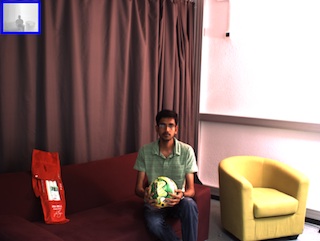}\label{fig:res_diff}}\hfil
  \subfigure[The proposed method delivers a high-resolution depth map.]{\includegraphics[height=0.36\linewidth]{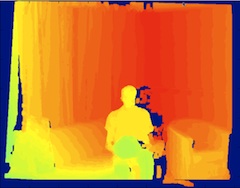}}  
  \caption{\footnotesize{(a) Two high-resolution color cameras (2.0MP at 30FPS) are combined with a single low-resolution time-of-flight camera (0.03MP at 30FPS). (b) A $144\times 177$ TOF image and a $1224\times 1624$ color image are shown at the true scale. (c) The depth map obtained with our method.
  The technology used by both these camera types allows simultaneous range and photometric data acquisition with an extremely accurate temporal synchronization, which may not be the case with other types of range cameras such as the current version of~Kinect. }}	
  \label{fig:camera_triplet}`
\end{figure}

Alternatively, \textit{active-light} range sensors, such as time-of-flight (TOF) or structured-light cameras, can be used to directly measure the 3D structure of a scene at video frame-rates.  However, the spatial resolution of currently available range sensors is lower than high-definition (HD) color cameras, the luminance sensitivity is poorer and the depth range is limited. The range-sensor data are often noisy and incomplete over extremely scattering parts of the scene, e.g., non-Lambertian surfaces. Therefore it is not judicious to rely solely on range-sensor estimates for obtaining 3D maps of complete scenes. Nevertheless, range cameras provide good initial estimates independently of whether the scene is textured or not, which is not the case with stereo matching algorithms. These considerations show that it is useful to combine the active-range and the passive-parallax approaches, in a \textit{mixed} system.  Such a system can overcome the limitations of both the active- and passive-range (stereo) approaches, when considered separately, and provides accurate and fast 3D reconstruction of a scene at high resolution, e.g. $1200\times 1600$ pixels at 30 frames/second, as in Fig.~\ref{fig:camera_triplet}. 

% Probelm statement end

% Related work

\subsection{Related Work} The combination of a depth sensor with a color camera has been exploited in several robotic applications such as object recognition~\cite{Gould:M2SFA208, Stuckler-2010, Attamimi-IROS-2010}, person awareness, gesture recognition~\cite{Droeschel-2011}, simultaneous localization and mapping (SLAM)~\cite{schwarz2011wmvc, Jebari-2011}, robotized plant-growth measurement~\cite{Alenya-2011}, etc.
These methods have to deal with the difficulty of noise in depth measurement and the inferior resolution of range data as compared to the color data. Also, most of these systems are limited to RGB-D, i.e., a \textit{single} color image combined with the range data. Interestingly enough, the recently commercialized Kinect\footnote{http://www.xbox.com/en-US/kinect} camera falls in the RGB-D family of sensors.
We believe that extending this model to an RGB-D-RGB sensor is extremely advantageous because it can incorporate stereoscopic matching and hence better deal with the problems mentioned above. 

Stereo matching has been one of the most studied paradigms in computer vision. Several papers, e.g., \cite{Scharstein,Seitz} present an overview of existing techniques and highlight recent progress in stereo matching and stereo reconstruction.  Algorithms based on a greedy local search are typically fast but frequently fail to reconstruct the poorly textured regions or ambiguous surfaces. Global methods formulate the matching task as a single optimization problem which leads to minimization of an Markov random field (MRF) energy function of the image similarity likelihood and a prior on the surface smoothness. These algorithms solve some of the aforementioned problems of local methods but are very complex and computationally expensive since optimizing an MRF-based energy function is an NP-hard problem in the general case.

A practical tradeoff between the local and the global methods in stereo is the seed growing class of algorithms~\cite{Cech-PAMI-2010,Cech-CVPR-2011,Cech-CVPR-2007}. The correspondences are grown from a small set of initial correspondence seeds. Interestingly, they are not particularly sensitive to wrong input seeds.
%One of the most interesting built-in properties of these algorithms is that they are robust to a fair pecentage of wrong initial correspondences. 
They are significantly faster than the global approaches, but they have difficulties in presence of non textured surfaces; Moreover, in these cases they yield depth maps which are relatively sparse. Denser maps can be obtained by relaxing the matching threshold but this leads to erroneous growth, which is a natural tradeoff between the accuracy and density of the solution. Some form of regularization is necessary in order to take full advantage of these methods. 

%Prior based models based models have long been used for segmentation and stereo applications in computer vision ~\cite{Boykov2001, Belhumeur1992}. Prior based approaches acted with a likelihood model by incurring a additional cost with incoherence in the data in the energy minimization framework.

%Most of the initial simple prior based approaches acted on the simple likelihood model by incurring a additional cost with incoherence in the data in the energy minimization framework. As the area progressed, the clear separation between the likelihood and the prior term was diminished by the introduction of prior models conditioned on the data terms ~\cite{Lafferty2001}. 

Recently an external prior-based generative probabilistic model for stereo matching was proposed in~\cite{Geiger,Newcombe-CVPR-2010} for reducing the matching ambiguities. The prior used was based on surface-triangulation on initially-matched distinctive interest points in the images. Again, in the absence of textured regions, such support points are either not available or are not reliable enough and the priors are erroneous. Consequently, the methods produce artifacts in cases the priors win over the data and the solution is biased towards the incorrect priors. This clearly shows the need for more accurate prior models. \cite{Wang-CVPR-2011} integrates a regularization term based on the depth values of initially matched \textit{ground control points} in a global energy minimization framework. The ground control points are gathered using an accurate laser scanner. A laser scanner is difficult to operate and cannot provide range information fast enough such that it can be used in a practical robotic application.

A TOF camera is based on an active sensor principle\footnote{http://www.mesa-imaging.ch} that allows 3D data acquisition at video frame rates, e.g., 30FPS as well as accurate synchronization with any number of color cameras\footnote{http://www.4dviews.com}. A modulated near infrared light from the camera's internal lighting source is reflected by objects in the scene and travels back to the sensor, where its precise time of flight is measured independently at each of the sensor's pixel by calculating the phase delay between the emitted and the detected wave. A complete depth map of the scene can be obtained using this sensor at the cost of very low spatial resolution and coarse depth accuracy. %Moreover, the luminance sensitivity is poor and the depth range is limited.

The fusion of TOF data with stereo data has been recently studied. For example,~\cite{DalMutto3DPVT10} obtained a higher quality depth map, by a probabilistic ad-hoc fusion of TOF and stereo data. Work in~\cite{Zhu} merges the depth probability distribution function obtained from TOF and stereo. However both these methods are meant for improvement over the initial data gathered with the TOF camera and the final depth-map result is still limited to the resolution of the TOF sensor. The method proposed in this paper increases the resolution from 0.03MP to the full resolution of the color cameras being used, e.g., 2MP.

The problem of depth map upsampling has been previously addressed. In~\cite{Chan-2008} a noise-aware filter for adaptive multi-lateral upsampling of TOF depth maps is presented. The work described in~\cite{Gould:M2SFA208} extends the model of~\cite{Diebel05b} and demonstrates that the object detection accuracy can be significantly improved by combining a state-of-art 2D object detector with 3D depth cues. The approach deals with the problem of resolution mismatch between range- and color-data using an MRF-based super-resolution technique in order to infer the depth at every pixel. The proposed method is slow: It takes around 10 seconds to produce a $320 \times 240$ depth image. All of these methods are limited to depth-map upsampling using only a single color image and do not exploit the added advantage offered by stereo matching, which can highly enhance the depth map both qualitatively and quantitatively. Recently, \cite{Fischer_ICRA_2011} proposed a method which combines TOF estimates  with stereo in a semiglobal matching framework. However, at pixels where TOF disparity estimates are available, the image similarity term is ignored. This make the method quite susceptible to errors in regions where TOF estimates are not precise, especially in textured regions where stereo itself is reliable.

% TOF with a stereo pair using a semiglobal optimization framework but it fails to make the full advantage of the provided TOF data.}

% Main contributions

\subsection{Contributions} 

%The sensor setup is shown in Fig.~\ref{fig:camera_triplet}. We combine 2 HD color images with low resolutions TOF sensor (RGB-D-RGB) to yield a high definition depth image (of the same resolution as the RGB cameras) in contrast to RGB-D based systems to exploit the added advantage offered by the stereo algorithms.

In this paper we propose a novel Bayesian method for incorporating range data within a robust seed-growing algorithm for stereoscopic matching~\cite{Cech-PAMI-2010}. A calibrated system composed of an active range sensor and a stereoscopic color-camera pair \cite{Miles}, e.g., Fig.~\ref{fig:camera_triplet}, allows the range data to be projected onto each one of the two images, thus providing an initial sparse set of point-to-point correspondences (seeds) between the two images. This initial seed set is used in conjunction with the seed-growing algorithm proposed in \cite{Cech-PAMI-2010}. The novelty is that the range data are used as the vertices of a mesh-based surface representation which, in turn, is used as a prior to regularize the image-based matching procedure. The Bayesian \textit{fusion}, between the mesh-based surface (initialized from the sparse range data) and the seed-growing stereo matching algorithm itself, combines the merits of the two 3D sensing methods and overcomes the limitations outlined above. 
The proposed fusion model can be incorporated within virtually any stereo algorithm that is based on energy minimization and which requires proper initialization. It is, however, particularly efficient and accurate when used in combination with match-propagation methods.

% Organization

The remainder of this paper is structured as follows: Section~\ref{section:method}  describes the proposed range-stereo fusion algorithm. 
Experimental results on a real dataset and evaluation of the method, are presented in section~\ref{section:evaluation}. Finally, section~\ref{section:conclusions} draws some conclusions.% and discusses directions for future work.

%% file: method.tex
%\begin{figure}[t]
%\centering
%\includegraphics[width=0.5\linewidth]{res1.png} 
%\caption{\footnotesize Resolution difference in TOF sensor depth image and full resolution color image (TOF depth image is shown in inset at left top corner)}
%\label{fig:res_diff}
%\end{figure}

As outlined above, the TOF camera provides a low-resolution depth map of a scene. This map can be projected onto the left and right images associated with the stereoscopic pair, using the projection matrices estimated by the calibration method described in~\cite{Miles}.
Projecting a single 3D point $(x,y,z)$ gathered by the TOF camera onto the \textit{rectified} images provides us with a pair of corresponding points $(u,v)$ and $(u',v)$ in the respective images. Each element $(u,u',v)$ denotes a point in the disparity space{\footnote{The disparity space is a space of all potential correspondences~\cite{Scharstein}.}. Hence, projecting all the points obtained with the TOF camera gives us a sparse set of 2D point correspondences. This set is termed as the set of initial support points or \textit{TOF seeds}. 

These initial support points are used in a variant of the seed-growing stereo algorithm~\cite{Cech-PAMI-2010, Cech-CVPR-2007} which further grows them into a denser and higher resolution disparity map. The seed-growing stereo algorithms propagate the correspondences by searching in the small neighborhood of given seed correspondences. Hereby, only a small fraction of disparity space is visited, which makes the algorithm extremely efficient from a computational point of view. The limited neighborhood also gives a kind of implicit regularization, nevertheless the solution can be arbitrarily complex, since multiple seeds are provided. 

The integration of range data within the seed-growing algorithm required two major modifications: (1)~The algorithm is using TOF seeds instead of the seeds obtained by matching distinctive image features, such as interest points, between the two images, and (2)~the growing procedure is regularized using a similarity statistic which takes into account  the photometric consistency as well as the depth likelihood based on disparity estimate by interpolating the rough triangulated TOF surface. This can be viewed as a prior cast over the disparity space. 

\begin{algorithm}[t]
\algsetup{linenosize=\scriptsize} 
\caption{Growing algorithm with sensor fusion}
\label{al:Alg1}
\begin{algorithmic}[1]
\REQUIRE Rectified images ($I_L, I_R$), \\
initial correspondence seeds $\cal S$, \\
image similarity threshold $\tau$.\\[1ex]
\STATE Compute the prior disparity map $D_p$ by interpolating seeds $\cal S$.\label{s:Dp} 
\STATE Compute $\simil(s | I_L,I_R,D_p)$ for each every seed $s \in \cal S$.\label{s:corr}
\STATE Initialize empty disparity map $D$ of size $I_L$ (and $D_p$).
\REPEAT
\STATE Draw seed $s \in \cal S$ of the best $\simil(s | I_L,I_R,D_p)$ value.\label{s:best}
\FOR{each of the four best neighbors  $i\! \in\! \{1,2,3,4\}$\\
~$q_i^\ast = (u, u', v)$ =  $\underset{q \in {\cal N}_i(s)}{\operatorname{argmax}}$  $\simil(q | I_L,I_R,D_p)$\\\hspace{-1ex}} \label{s:for}
\STATE $c:=\simil(q_i^\ast | I_L,I_R,D_p)$
\IF{$c \ge \tau$  {\bf and} pixels not matched yet} \label{s:if}
\STATE Update the seed queue ${\cal S} := {\cal S} \cup \{ q_i^\ast \}$.
\STATE Update the output map ${D}(u,v) = u-u'$.
\ENDIF
\ENDFOR
\UNTIL{$\cal S$ is empty}
\RETURN disparity map $D$.
\end{algorithmic}
\end{algorithm}

The growing algorithm is summarized in Sec.~\ref{sec:growing}. The processing of the TOF correspondence seeds is explained in Sec.~\ref{sec:seeds}, and the sensor fusion based similarity statistic is described in Sec.~\ref{sec:fusion}.

\subsection{The Growing Procedure} \label{sec:growing}
The growing algorithm is sketched in pseudo-code as Alg.~\ref{al:Alg1}. The input is a pair of rectified images $(I_L, I_R)$, a set of (refined) TOF seeds $\cal S$, and a parameter $\tau$ which directly controls a trade-off  between accuracy and density of the matching. The output is a disparity map $D$ which relates pixel correspondences between the input images.

First, the algorithm computes the prior disparity map $D_p$ by interpolating TOF seeds. Map $D_p$ is of the same size as the input images and the output disparity map, Step~\ref{s:Dp}. Then, a similarity statistic $\simil(s | I_L,I_R,D_p)$ of the correspondence, which measures both the photometric consistency of the possible correspondence and consistency with the prior, is computed for all seeds $s = (u,u',v) \in \cal S$, Step~\ref{s:corr}. Recall that the seed $s$ stands for a correspondence $(u,v)\leftrightarrow (u',v)$ between pixels in the left and the right images. 
For each seed, the algorithm searches other correspondences in the surroundings of the seeds by maximizing the similarity statistic. This is done in a 4-neighborhood $\{{\cal N}_1, {\cal N}_2, {\cal N}_3. {\cal N}_4\}$ of the pixel correspondence, such that in each respective direction (left, right, up, down) the algorithm searches the disparity in a range $\pm 1$ pixel from the disparity of the seed, Step~\ref{s:for}. If the similarity statistic of a candidate exceeds threshold $\tau$, then a new correspondence is found, Step~\ref{s:if}. It becomes a new seed, and output disparity map $D$ is updated. The process repeats until there are no more seeds to be grown. 

{The algorithm is fairly insensitive to wrong initial seeds. Since the seeds compete to be matched in the best first strategy, the wrong seeds typically have low score $\simil(s)$ and therefore when they are drawn in Step~\ref{s:best}, the involved pixels are likely to have been matched already.}
For more details on the growing algorithm, we refer to~\cite{Cech-CVPR-2007,Cech-PAMI-2010}.

\subsection{TOF Seeds and Their Refinement} \label{sec:seeds}
The original version of the seed-growing stereo algorithm~\cite{Cech-CVPR-2007} uses an initial set of seeds $\cal S$ obtained by detecting interest points in both images and matching them. Here, we propose to use TOF seeds. As already outlined, these seeds are obtained by projecting the low-resolution depth map associated with the TOF camera onto the high-resolution images. Likewise in the case of interest points, this yields a sparse set of seeds, e.g., approximately 25,000 seeds in the case of the TOF camera used in our experiments. Nevertheless, one of the main advantages of the TOF seeds over the interest points is that they are regularly distributed across the images regardless of the presence/absence of texture. This is not the case with interest points whose distribution strongly depends on texture as well as lighting conditions, etc. Obviously, regularly distributed seeds will provide a better coverage of the observed scene.

\begin{figure}
\begin{center}
\includegraphics[width=0.99\linewidth]{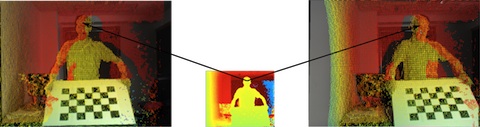} 
\end{center}
   \caption{{{Projection of TOF sensor data on left and right images. The points are color coded and the color represents disparity such that colder colors are closer to the cameras. The images are not in the true scale. Notice wrong correspondences on the computer screen due to low reflectance and artifacts along occlusion boundaries.} }}
\label{fig:tof_reprojection}
\end{figure}

\begin{figure}%[t]
  \centering
  \includegraphics[width=0.42\textwidth]{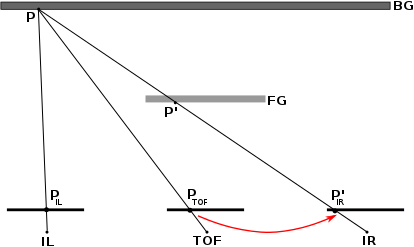}
  \caption{\footnotesize{The effect of occlusions. A background (BG) point $P$ is seen in the left image (IL) and in the TOF image, while it is occluded by a foreground object (FG) and hence not seen in the right image (IR). In the process of reprojection of 3D TOF points, a wrong correspondence ($P_{IL} \leftrightarrow P'_{IR}$) is produced. }}
\label{fig:projection_problem}
\end{figure}

However, TOF seeds are not always accurate. Indeed, when a 3D point is projected onto the left- and the right-image, it does not always yield  a valid stereo match. 
There may be several sources of error which make the TOF seeds less reliable than one would have expected, e.g., Fig.~\ref{fig:tof_reprojection} and Fig.~\ref{fig:projection_problem}. In detail:
\begin{enumerate}\itemsep=0pt

\item \textit{Imprecision due to the calibration data}. The transformations allowing to project the 3D TOF points onto the 2D images are obtained via a complex sensor calibration process \cite{Miles}. This introduces a localization error up to 2-3 pixels.

\item \textit{Outliers due to the physical/geometric properties of the scene.} Range sensors are based on active light and on the assumption that the active beam of light travels from the sensor and back to it. There are a number of situations where the beam is lost, such as specular surfaces, absorbing surfaces (such as fabric), scattering surfaces (such as hair), slanted surfaces, bright surfaces (computer monitors), faraway surfaces (limited range), or when the beam travels in an unpredictable way, such a multiple reflections. 

\item \textit{The TOF- and 2D cameras observe the scene from slightly different points of view}. Therefore, it may occur that a 3D point that is present in the TOF image is only seen into the left or right image, e.g., Fig.~\ref{fig:projection_problem}.

\end{enumerate} 

Therefore, a fair percentage of the TOF seeds are \textit{outliers}. Although the seed-growing stereo matching algorithm is robust to the presence of outliers in the initial set of seeds, as already explained in section~\ref{sec:growing},
we implemented a straightforward refinement step in order to detect and eliminate these kind of bad seed data, prior to applying Alg.~\ref{al:Alg1}. Firstly, the seeds that lie in low-intensity (very dark) regions are discarded since TOF-based range data are not reliable in these cases. Secondly, in order to handle the background-to-foreground occlusion effect just outlined, 
we detect seeds which are not uniformly distributed across image regions. Indeed, projected 3D points lying on smooth fronto-parallel surfaces form a regular image pattern of seeds, while projected 3D points that belong to a background surface and which project onto a foreground image region do not form a regular pattern. See occlusion boundary in Fig.~\ref{fig:orig_seeds}.

{Such seeds are detected by counting the occupancy of small $5 \times 5$ pixel windows around every seed point in both images. If there is more than one seed point, the seeds belonging to the background are discarded. Refined set of seeds is shown in Fig.~\ref{fig:refined_seeds}. The refinement procedure typically filters 10-15\% of all seed points.}
%Explain here what is done in practice and illustrate with Fig.~\ref{fig:refined_seeds} and Fig.~\ref{fig:orig_seeds}}.

\begin{figure}[t]
  \centering
\subfigure[Original set of seeds] {\label{fig:orig_seeds}\includegraphics[width=0.22\textwidth]{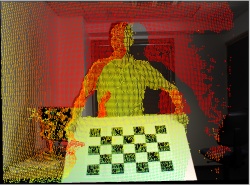}} \hspace{2.0mm}
\subfigure[Refined set of seeds]{\label{fig:refined_seeds}\includegraphics[width=0.22\textwidth]{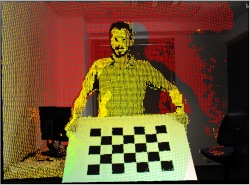}} 
\caption{{\footnotesize An example of the effect of correcting the set of seeds on the basis that they should be regularly distributed.}}

\end{figure}

\begin{figure}[t]
  \centering
\subfigure[Delaunay Triangulation on original seeds] {\label{fig:tri1}\includegraphics[width=0.22\textwidth]{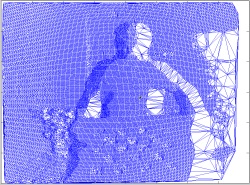}} \hspace{2.0mm}
\subfigure[Delaunay Triangulation on refined seeds]{\label{fig:tri2}\includegraphics[width=0.22\textwidth]{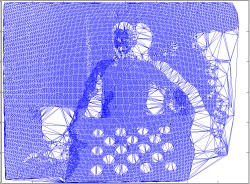}} 

\subfigure[Prior obtained on original seeds]{\label{fig:prior1}\includegraphics[width=0.22\textwidth]{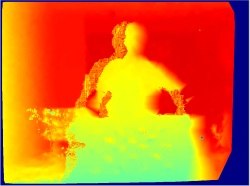}}\hspace{2.0mm}
\subfigure[Prior obtained on refined seeds]{\label{fig:prior2}\includegraphics[width=0.22\textwidth]{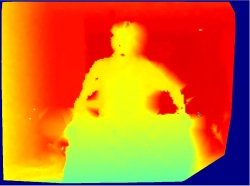}}
 
\caption{Triangulation and prior disparity map $D_p$. These are shown using both raw seeds (a), (c) and refined seeds (b), (d). A positive impact of the refinement procedure is clearly visible.}
\label{fig:triangulation}
\end{figure}

%a simple idea was incorporated to compensate for the overlapping background points over foreground. Projection of TOF sensor data on the color images gives a regular pattern of projected seeds due to the large resolution differences. Whenever an overlap of background point over foreground occurs this regular pattern is disturbed. By analyzing the violation in this regular pattern the ambiguous points can be identified. The correct disparity value of these identified points was approximated using the information of the neighbouring projected points. 

% Presence of seeds in every component of disparity space, i.e.\ in every contiguous 3D surface, enhances the performance of the correspondence growing algorithm.

\subsection{Similarity Statistic Based on Sensor Fusion} \label{sec:fusion}
The original seed-growing matching algorithm~\cite{Cech-CVPR-2007} uses Moravec's normalized cross correalation~\cite{Moravec-77} (MNCC), 
%namely:
\begin{equation}
\mbox{simil}(s) = \mbox{MNCC}(w_L,w_R) = \dfrac{2 \mbox{cov}(w_L,w_R)}{\mbox{var}(w_L)+ \mbox{var}(w_R) + \epsilon }
\label{eq:MNCC1}
\end{equation}
as the similarity statistic to measure the photometric consistency of a correspondence $s : (u,v)\leftrightarrow (u',v)$. We denote by $w_L$ and $w_R$ the feature vectors which collect image intensities in small windows of size $n \times n$ pixels centered at $(u,v)$ and $(u'v)$ in the left and right image respectively. The parameter $\epsilon$ prevents instability of the statistic in cases of low intensity variance. {This is set as the machine floating point epsilon}. The statistic has low response in textureless regions and therefore the growing algorithm does not propagate the correspondences across these regions. Since the TOF sensor can provide seeds without the presence of any texture, we propose a novel similarity statistic,  $\simil(s|I_L,I_R,D_p)$. This similarity measure uses a different score for photometric consistency as well as an initial high-resolution disparity map $D_p$, both incorporated into the Bayesian model explained in detail below. 

The initial disparity map $D_p$ is computed as follows. A 3D meshed surface is built from a 2D triangulation applied to the TOF image. The disparity map $D_p$ is obtained via interpolation from this surface such that it has the same (high) resolution as of the left and right images. Fig.~\ref{fig:tri1} and \ref{fig:tri2} show the meshed surface projected onto the left high-resolution image and built from the TOF data, before and after the seed refinement step, which makes the $D_p$ map more precise. %\red{Notice that $D_p$ is imprecise especially around object boundaries and fine structures. Therefore, we use it together with a photometric consistency likelihood in joint formula.}

%The map $D_p$ is of the same resolution as the input images (and also the output disparity map). Due to low resolution of TOF depth map and after the refinement step described above, sparse seed points are projected into the images, i.e.\ for certain pixels the disparity is given. \red{For the rest of pixels, the prior disparity is estimated by interpolation using Delaunay triangulation. The 2D Delaunay triangulation defines a mesh, see Fig.~\ref{fig:tri1},~\ref{fig:tri1}. Each query pixel is assigned a disparity value of linear interpolation within the triangle containing the query pixel.}
%
%Obviously, such an approximation is erroneous in occlusions and near object boundaries, see Fig.~\ref{fig:prior1},~\ref{fig:prior2}, but can serve as a good prior to regularize the growing procedure in ambiguous regions. 
Let us now consider the task of finding an optimal high-resolution disparity map. For each correspondence $(u,v)\leftrightarrow (u',v)$ and associated disparity $d = u - u'$ we seek an optimal disparity $d^\ast$ such that:
\begin{equation}
 d^\ast = \underset{d}{\argmax} ~ P(d | I_L, I_R, D_p).
\end{equation}
By applying the Bayes' rule, neglecting constant terms, assuming that the distribution  $P(d)$ is uniform in a local neighborhood where it is sought (Step.~\ref{s:for}), and considering conditional independence $P(I_l,I_r,D | d) = P(I_L,I_R | d) P(D_p|d)$, we obtain:
\begin{equation}
 d^\ast = \underset{d}{\argmax} ~ P(I_L,I_R | d) ~ P(D_p|d),
\end{equation}
where the first term is the image likelihood and the second term is the range-sensor likelihood. 
We define the  image and range-sensor likelihoods as:
\begin{multline}
P(I_L, I_R|d) \propto \mbox{EXPSSD}(w_{L},w_{R}) = \\ =\mbox{exp}\left(- \frac{\sum_{i=1}^{n\times n} (w_{L}(i) - w_{R}(i))^{2}}{\sigma_{s}^2 \sum_{i=1}^{n\times n} (w_{L}(i)^{2} + w_{R}(i)^{2} )}\right), \label{eq:expssd}
\end{multline}
and as
\begin{equation}
P(D_{p}|d) \propto \mbox{exp}\left(- \frac{(d-d_{p})^{2}}{2 \sigma_{p}^{2}}\right)
\end{equation}
respectively, where $\sigma_{s}$ are $\sigma_{p}$ two normalization parameters. 
Therefore, the new similarity statistic becomes
\begin{multline}
\hspace{-2ex}\simil(s|I_L,I_R,D_p) = \mbox{EPC}(w_L,w_R,D_p) = \\ = \mbox{exp} \left(- \frac{\sum_{i=1}^{n\times n} (w_{L}(i) - w_{R}(i))^{2}}{\sigma_{s}^2 \sum_{i=1}^{n\times n}( w_{L}(i)^{2} + w_{R}(i)^{2} ) } - \frac{(d-d_{p})^{2}}{2  \sigma_{p}^{2}}\right). \label{eq:EPC}
\end{multline}

Notice that the proposed image likelihood has a high response for correspondences associated with textureless regions. However, in such regions, all possible matches have similar image likelihoods. The proposed range-sensor likelihood regularizes the solution and forces it towards the one closest to the prior disparity map $D_p$. A tradeoff between these two terms can be obtained by tuning the parameters $\sigma_s$ and $\sigma_p$. 

%In the following section, we will demonstrate the proposed algorithm in several experiments. 

%\clearpage

%%%%%%%%%%%%%%%%%%%%%%%%%%%%%%%%%%%%%%%%%%%%%%%%%%%%%%%%%%%%%%%%%%%%%%%%%%%%%%%%%%%%%%%%%%%%%%%%%%%%%%%%%%%%%%%%%%%%%
%%%%%%%%%%%%%%%%%%%%%%%%%%%%%%%%%%%%%%%%%%%%%%%%%%%%%%%%%%%%%%%%%%%%%%%%%%%%%%%%%%%%%%%%%%%%%%%%%%%%%%%%%%%%%%%%%%%%%
%%%%%%%%%%%%%%%%%%%%%%%%%%%%%%%%%%%%%%%%%%%%%%%%%%%%%%%%%%%%%%%%%%%%%%%%%%%%%%%%%%%%%%%%%%%%%%%%%%%%%%%%%%%%%%%%%%%%%

%% file: experiments.tex
%\subsection{Data Acquisition and Calibration}

Our experimental setup comprises one Mesa Imaging $\mbox{Swissranger}^{\mbox{\tiny TM}}$ SR4000 TOF camera and a pair of high-resolution Point~Grey\footnote{http://www.ptgrey.com/} color cameras, e.g., Fig.~\ref{fig:camera_triplet}. The two color cameras are mounted on a rail with a baseline of about 49 cm and the TOF camera is approximately midway between them. All three optical axes are approximately parallel. The resolution of the TOF image is of $144 \times 177$ and the color cameras have a resolution of $1224 \times 1624$. Recall that Fig.~\ref{fig:res_diff} highlights the resolution differences between the TOF and color images.This camera system was calibrated using the method described in~\cite{Miles}. 

%A checkerboard with $8cm \times 8cm$ square was used as the calibration object. A single capture gives us an image triplet (left image, right image, TOF image) with 35 vertices in full subpixel correspondence. Several captures were made to obtain a set of at least 25-30 valid image triplets. %\red{The images were radially undistorted.}

%Paper~\cite{Miles} proposed a new method for geometric mapping between Range and Parallax data without requiring an intermediate calibration of the binocular system. A point based alignment technique proposed in the paper can be used to obtain the projective alignment of Range and Parallax data given a set of point-correspondences between the range and the color cameras. Such a geometric mapping enables us to re-project the range data into the binocular cameras, which makes it possible to associate color and texture with each point in the range representation~\cite{Miles}.

In all our experiments, we set the parameters of the method as follows: Windows of $5\times5$ pixels were used for matching ($n = 5$), matching threshold in Alg.~1 to $\tau = 0.5$, the balance between the photometric and range sensor likelihoods was set to $\sigma_s^2 = 0.1$ and to $\sigma_p^2 = 0.001$ in~(\ref{eq:EPC}).

We show both qualitatively and quantitatively (using datasets with ground-truth) benefits of the range sensor and an impact of particular variants of the proposed fusion model integrated in the growing algorithm. Namely, we compare results of (i)~the original stereo algorithm~\cite{Cech-CVPR-2007} with MNCC correlation and Harris seeds (MNCC-Harris), (ii)~the same algorithm with TOF seeds (MNCC-TOF), (iii)~the algorithm which uses EXPSSD similarity statistic instead with both Harris (EXPSSD-Harris) and TOF seeds (EXPSSD-TOF), and (iv)~the full sensor fusion model of the regularized growth (EPC). Finally small gaps of unassigned disparity in the disparity maps were filled by a primitive procedure which assigns median disparity in the $5 \times 5$ window around the gap (EPC - gaps filled). These small gaps  usually occur in slanted surfaces, since Alg.~1 in Step.~\ref{s:if} enforces one-to-one pixel matching. Nevertheless this way, they can be filled easily if needed.

%First, we demonstrate the benefits of replacing original seeds given by Harris prematcher with TOF points as seeds in the original Seed Growing Algorithm. Then, we show a significant improvement of the results by using the algorithm with the full Bayesian model including the prior term. The depth maps obtained using different appraoches have been compared on a real world dataset. Then, we compare the surface reconstructions using the Bayesian model with direct triangulation based reconstruction from initial TOF seeds. 

%Our code combines Matlab and C++ using mex compiler. Disparity maps are shown in color: colder colors codes smaller disparities and warmer color are larger disparities, dark blue areas with unassigned disparities usually due to occlusions.  

\subsection{Real-Data Experiments}

\begin{figure}[t]

  \centering

\def\hght{61pt}

\subfigure[Input data RGB-TOF-RGB (the true scale is shown in Fig.~\ref{fig:res_diff}). ] {\label{fig:63_1} \includegraphics[height=\hght]{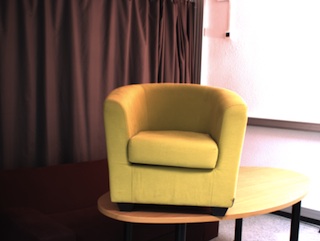} \parbox[b]{1.9cm}{\hfil\includegraphics[height=0.6cm]{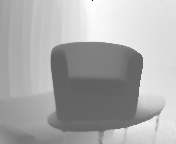}\\[2ex]~\hfil} \includegraphics[height=\hght]{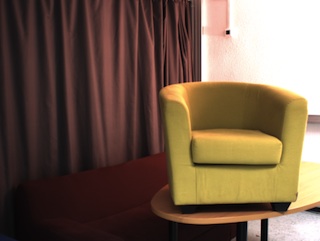}}

\subfigure[MNCC-Harris] {\label{fig:63_mh} \includegraphics[height=\hght]{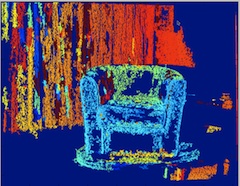}}\hspace{-1ex} 
\subfigure[MNCC-TOF] {\label{fig:63_mt} \includegraphics[height=\hght]{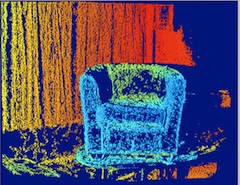}}\hspace{-1ex} 
\subfigure[EXPSSD-Harris] {\label{fig:63_eh} \includegraphics[height=\hght]{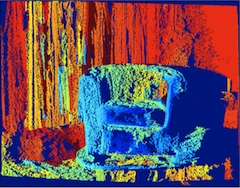}} \\

\subfigure[EXPSSD-TOF] {\label{fig:63_et} \includegraphics[height=\hght]{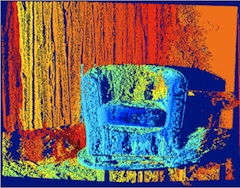}}\hspace{-1ex} 
\subfigure[{\bf EPC} (proposed)] {\label{fig:63_epc} \includegraphics[height=\hght]{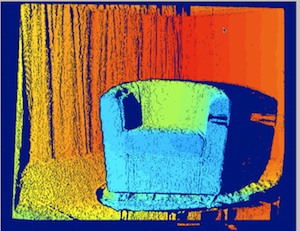}}\hspace{-1ex} 
\subfigure[{\bf EPC} (gaps filled)] {\label{fig:63_epc_filled} \includegraphics[height=\hght]{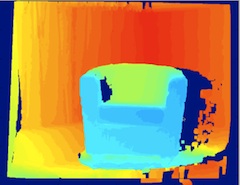}} 
 
\caption{ SET-1.  (a) A triplet composed of a pair of color images (left and right) and a TOF image (middle), results obtained (b) using the seed growing stereo algorithm~\cite{Cech-CVPR-2007} combined Harris seeds and MNCC statistic, (c) using TOF seeds and MNCC statistic , (d) using Harris seeds and EXPSSD statistic, (e) using TOF seeds with EXPSSD statistics. Results obtained with proposed full stereo-TOF fusion model using EPC similarity statistic (f) and full model EPC after filling small gaps (g).}

\label{fig:result_set1}
\end{figure}

\begin{figure}[t]
\centering
\def\hght{61pt}
\subfigure[Input data RGB-TOF-RGB (the true scale is shown in Fig.~\ref{fig:res_diff}).] {\label{fig:69_1} \includegraphics[height=\hght]{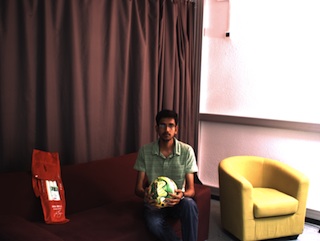} \parbox[b]{1.9cm}{\hfil\includegraphics[height=0.6cm]{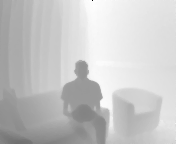}\\[2ex]~\hfil} \includegraphics[height=\hght]{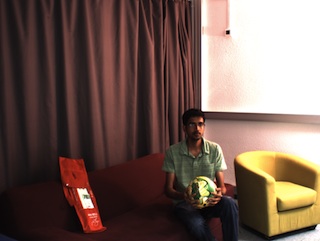}}

\subfigure[MNCC-Harris] {\label{fig:69_mh} \includegraphics[height=\hght]{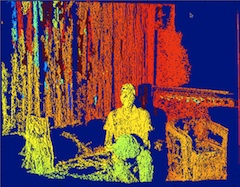}}\hspace{-1ex} 
\subfigure[MNCC-TOF] {\label{fig:69_mt} \includegraphics[height=\hght]{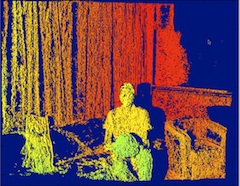}}\hspace{-1ex} 
\subfigure[EXPSSD-Harris] {\label{fig:69_eh} \includegraphics[height=\hght]{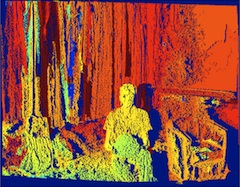}} 

\subfigure[EXPSSD-TOF] {\label{fig:69_et} \includegraphics[height=\hght]{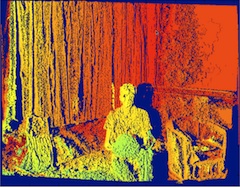}}\hspace{-1ex} 
\subfigure[{\bf EPC} (proposed)] {\label{fig:69_epc} \includegraphics[height=\hght]{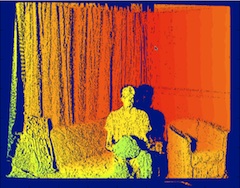}}\hspace{-1ex} 
\subfigure[{\bf EPC} (gaps filled)] {\label{fig:69_epc_filled} \includegraphics[height=\hght]{69_filled_dmap.jpg}} 
 
\caption{SET-2. Please refer to the caption of Fig.~\ref{fig:result_set1} for explanation.}

  \label{fig:result_set2}
\end{figure}

We captured two real-world datasets using the camera setup described above, SET-1 in Fig.~\ref{fig:result_set1} and SET-2 in \ref{fig:result_set2}. Notice that in both of these examples the scene surfaces are weakly textured.

\bigskip
\subsubsection{Comparisons between disparity maps}  
Results as disparity maps are shown color-coded, such that warmer colors are further away from the cameras and unmatched pixels are dark blue.

In Fig.~\ref{fig:63_mh}, we can see that the original algorithm~\cite{Cech-CVPR-2007} has difficulties in low textured regions which results in large unmatched regions due to MNCC statistic~(\ref{eq:MNCC1}), and it produces several mismatches over repetitive structures on the background curtain, due to erroneous (mismatched) Harris seeds. In Fig.~\ref{fig:63_mt}, we can see that after replacing the sparse erratic Harris seeds with uniformly distributed mostly correct TOF seeds, results improve significantly. There are no more mismatches on the background, but unmatched regions are still large. In Fig.~\ref{fig:63_eh}, the EXPSSD statistic (\ref{eq:expssd}) was used instead of MNCC which causes similar mismatches as in Fig.~\ref{fig:63_mh}, but unlike MNCC there are matches in textureless regions, nevertheless mostly erratic. The reason is that unlike MNCC statistic the EXPSSD statistic has high response in low textured regions. However, since all disparity candidates have equal (high) response inside such regions, the unregularized growth is random, and produces mismatches. The situation does not improve much using the TOF seeds, as shown in Fig.~\ref{fig:63_et}. Significantly better results are finally shown in Fig.~\ref{fig:63_epc} which is using the full proposed fusion model EPC~(\ref{eq:EPC}). The EPC statistic compared to EXPSSD has the additional regularizing range sensor likelihood term which guides the growth in ambiguous regions and attracts the solution towards the rough estimate of the TOF camera. Results are further refined by filling small gaps in Fig.~\ref{fig:63_epc_filled}. Similar observations can be made in Fig.~\ref{fig:result_set2}. The proposed model clearly outperforms the other discussed approaches. 

\bigskip
\subsubsection{Comparisons between reconstructed surfaces}
For the proper analysis of a stereo matching algorithm it is important to inspect the reconstructed 3D surfaces. Indeed the visualization of the disparity/depth maps can sometimes be misleading. Surface reconstruction reveals fine details in the quality of the results. This is in order to qualitatively show the gain of the high-resolution depth map produced by the proposed algorithm with respect to low-resolution depth map of the TOF sensor. 

\begin{figure}
\centering
\hspace{-1.5ex}\subfigure[Dense surface reconstruction using the disparity map $D_{p}$ corresponding to a 2D triangulation of the TOF data points. Zoomed sofa chair and zoomed T-shirt from SET-2 in Fig.~\ref{fig:69_1}.]
{\label{fig:69_t1} \includegraphics[height=82pt]{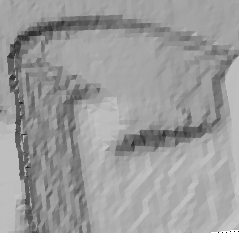}
\includegraphics[height=82pt]{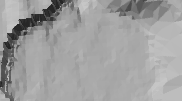}} 

\subfigure[Surface reconstruction using the proposed algorithm (EPC) shown on the same zoomed areas as above, i.e., Fig.~\ref{fig:69_epc_filled}.]
 {\label{fig:69_r1}\includegraphics[height=82pt]{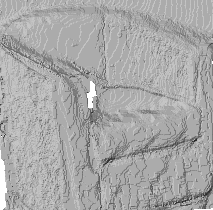}   
\includegraphics[height=82pt]{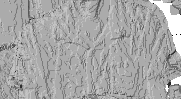}} 

\caption{The reconstructed surfaces are shown as relighted 3D meshed for (a) the prior disparity map $D_{p}$ (2D triangulation on projected and refined TOF seeds), and (b) for the disparity map obtained using the proposed algorithm. Notice the fine surface details which were recovered by the proposed method.}
\label{fig:result_tex}
\end{figure}

In order to provide a fair comparison, we show the reconstructed surfaces associated with the \textit{dense} disparity maps $D_p$ obtained after 2D triangulation of the TOF data points,  Fig.~\ref{fig:69_t1}, as well as the reconstructed surfaces associated with the disparity map obtained with the proposed method, Fig.~\ref{fig:69_r1}. Clearly, much more of the surface details are recovered by the proposed method. Notice precise object boundaries and fine details, like a pillow on the sofa chair and a collar of the T-shirt, which appear in Fig.~\ref{fig:69_r1}. This qualitatively corroborates the precision of the proposed method compared to the TOF data. 

\subsection{Ground-Truth Evaluation}

To quantitatively demonstrate the validity of the proposed algorithm, we carried out an experiment on datasets with associated ground-truth results. Similarly to~\cite{DalMutto3DPVT10} we use Middlebury dataset~\cite{Scharstein} and simulated the TOF camera by sampling the ground-truth disparity map. 

We used the Middlebury-2006 dataset\footnote{http://vision.middlebury.edu/stereo/data/scenes2006/}. On purpose, we selected three challenging scenes with weakly textured surfaces: Lampshade-1, Monopoly, Plastic. The input images are of size $1330 \times 1110$ pixels. We took every 10th pixel in a regular grid to simulate the TOF camera. This gives us about 14k of TOF points, which is roughly the same ratio to color images as for the real sensors. We are aware that simulation TOF sensor this way is naive, since we do not simulate any noise or artifacts, but we believe that for validating the proposed method this is satisfactory. 

\begin{figure*}
\def\wdth{0.12\linewidth}
\def\f{\footnotesize}
\tabcolsep=1pt
\begin{tabular}{cccccccc}

{\footnotesize Left image} & {\footnotesize  MNCC-Harris} & {\footnotesize  MNCC-TOF} & {\footnotesize  EXPSSD-Harris} & {\footnotesize  EXPSSD-TOF} & {\footnotesize  {\bf EPC}} & {\footnotesize  {\bf EPC} (gaps filled)} & {\footnotesize  Ground-truth} \\[1ex]

\includegraphics[width=\wdth]{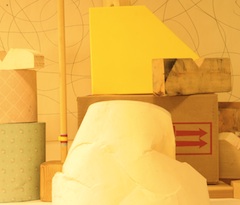}&
\includegraphics[width=\wdth]{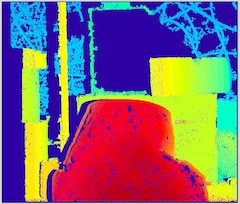}&
\includegraphics[width=\wdth]{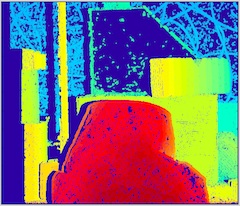}&
\includegraphics[width=\wdth]{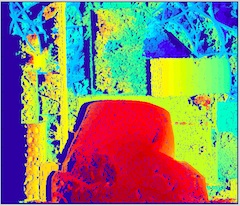}&
\includegraphics[width=\wdth]{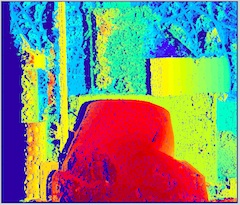}&
\includegraphics[width=\wdth]{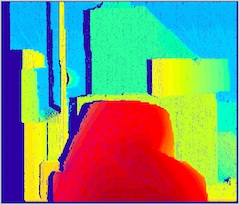}&
\includegraphics[width=\wdth]{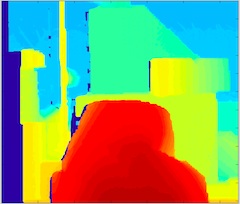} &
\includegraphics[width=\wdth]{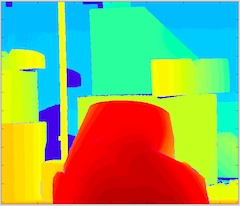} \\[-0.5ex]

Lampshade-1 & 61.5\% & 64.3\% & 44.9\% & 49.5\% & 88.8\% & 96.4\% & -- \\[1.5ex]

\includegraphics[width=\wdth]{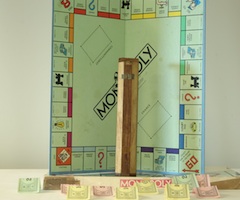}&
\includegraphics[width=\wdth]{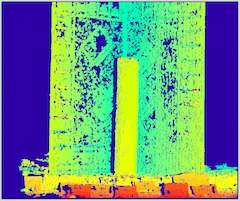}&
\includegraphics[width=\wdth]{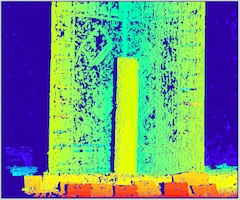}&
\includegraphics[width=\wdth]{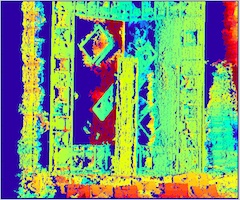}&
\includegraphics[width=\wdth]{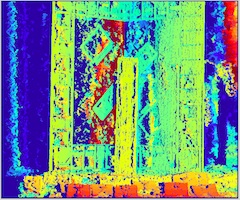}&
\includegraphics[width=\wdth]{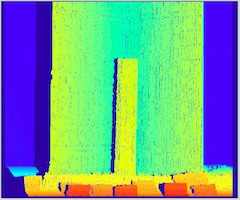}&
\includegraphics[width=\wdth]{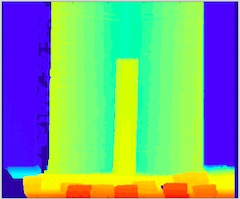} &
\includegraphics[width=\wdth]{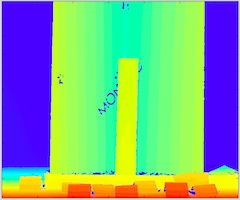} \\[-0.5ex]

Monopoly & 51.2\% & 53.4\% & 29.4\% & 32.1\% & 85.2\% & 95.3\% & -- \\[1.5ex]

\includegraphics[width=\wdth]{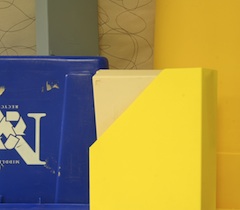}&
\includegraphics[width=\wdth]{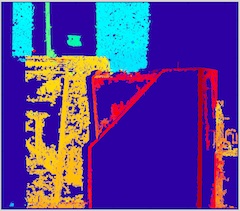}&
\includegraphics[width=\wdth]{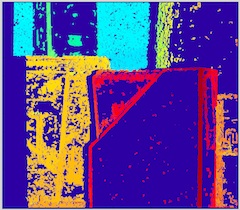}&
\includegraphics[width=\wdth]{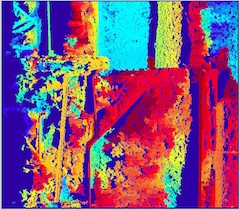}&
\includegraphics[width=\wdth]{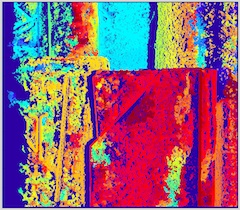}&
\includegraphics[width=\wdth]{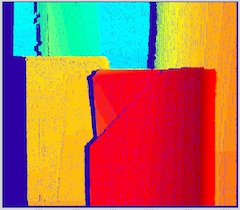}&
\includegraphics[width=\wdth]{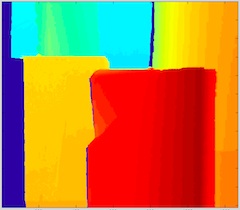} &
\includegraphics[width=\wdth]{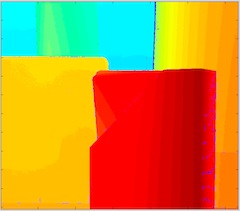} \\[-0.5ex]

Plastic & 25.2\% & 28.2\% & 13.5\% & 20.6\% & 88.7\% & 98.2\% & --

\end{tabular}
\caption{Middlebury dataset. We show left images, results of the same algorithms as in Fig.~\ref{fig:result_set1} and \ref{fig:result_set2}, and the ground-truth disparity maps. There are error statistics (percentage of correctly matched pixels) below the disparity maps. Observations from the real-world cases are confirmed quantitatively. The algorithm using the full model (EPC) gives clearly the best results.}
\label{fig:middlebury}
\end{figure*}

Results are shown in Fig.~\ref{fig:middlebury}. We show left input image, results of the same algorithms as in the previous section with the real sensor, and the ground-truth disparity map. For each disparity, we compute the percentage of correctly matched pixels in non-occluded regions. This error statistic is computed as number of pixels for which the estimated disparity differs from the ground-truth disparity by less than one pixel divided by number of all pixels in non-occluded regions. Notice that unmatched pixels are considered as errors of the same kind as mismatches. This is in order to allow a strict but fair comparison between algorithms which deliver solution of different density. The quantitative evaluation confirms the observation from the real-world setup. The proposed algorithm which uses the full sensor fusion model significantly outperforms all other tested variants. 

For the sake of completeness we also report error statistics for the prior disparity map $D_p$ which is computed by interpolating TOF seeds, see Step~\ref{s:Dp} of Alg.1. This is $92.9\%,~ 92.1\%,~ 96.0\%$ for Lampshade-1, Monopoly, Plastic scene respectively. These results are already quite good, which means the interpolation we use to construct the prior disparity map is appropriate. These scenes are mostly piecewise planar, which the interpolation captures well. On the other hand, recall that in the real case, not all the seeds are correct due to various artifacts of a range sensor. Nevertheless in all three scenes, the proposed algorithm (EPC with gaps filled) was able to further improve the precision up to $96.4\%,~95.3\%,~98.2\%$ for respective scenes. This is again consistent with the experiments with the real TOF sensor, where higher surface details were recovered, see Fig.~\ref{fig:result_tex}.

\subsection{Computational Costs}

The original growing algorithm~\cite{Cech-CVPR-2007} has low computational complexity due to intrinsic search space reduction. Assuming the input stereo images of size $n\times n$ pixels, the algorithm has the complexity of ${\cal O}(n^2)$, while any exhaustive algorithm has the complexity at least ${\cal O}(n^3)$,~\cite{Cech-CVPR-2011}. Factor $n^3$ is the size of the search space where the correspondences are sought, i.e.\ the disparity space. The growing algorithm does not compute similarity statistics of all possible correspondences, but efficiently traces out components of high similarity score around the seeds. This low complexity is beneficial especially for high resolution imagery, which allows precise surface reconstruction.

The proposed algorithm with all presented modifications does not represent any significant extra cost. Triangulation of TOF seeds and the prior disparity map computation is not really expensive as well as the computing the new EPC statistic instead of MNCC. 

For our experiments, we use an academic (combined Matlab and C) implementation which takes about {5 seconds} with 2 MP images. Nevertheless, recently~\cite{Dobias-ICCV-LDRMC-2011} presented an implementation of the growing algorithm~\cite{Cech-CVPR-2007} which runs in real-time in normal CPU, without parallel hardware. This indicates that a real-time implementation of the proposed algorithm would be feasible.

%% file: conclusion.tex
We have proposed a novel correspondence growing algorithm with fusion of a range sensor and a pair of passive color cameras to obtain accurate and dense 3D reconstruction of a given scene. The proposed algorithm is robust and performs well on both textured and texture less surfaces and ambiguous repetitive patterns. The algorithm exploits the strengths of TOF sensor and stereo matching between color cameras in a combined way to compensate for their individual weaknesses. The algorithm has shown promising results on difficult real world data as well as on challenging standard datasets which quantitatively corroborates its favourable properties. Together with the strong potential for real-time performance that we discussed, the algorithm would be practically very useful in many computer vision and robotic applications.